\documentclass[11pt,a4paper]{article}
\usepackage[hyperref]{emnlp2020}
\usepackage{times}
\usepackage{latexsym}
\usepackage{xspace}
\usepackage{graphicx}
\usepackage{multirow}
\usepackage{CJKutf8}

\newsavebox\FrameBox

\aclfinalcopy %
\def\aclpaperid{2539} %

\newcommand{\RN}[1]{%
  \textup{\uppercase\expandafter{\romannumeral#1}}%
}
\newcommand{\TC}[1]{\textcircled{\raisebox{-0.9pt}{#1}}}

\definecolor{dkgreen}{RGB}{0,130,0}

\newcommand{\wikinli}{\textsc{WikiNLI}\xspace}

\newcommand{\wordnet}{WordNet\xspace}
\newcommand{\wikidata}{Wikidata\xspace}
\newcommand{\wikipedia}{Wikipedia\xspace}

\newcommand{\bert}{BERT\xspace}
\newcommand{\roberta}{RoBERTa\xspace}
\newcommand{\robertalarge}{RoBERTa-large\xspace}

\newcommand{\bertlarge}{BERT-large\xspace}

\newcommand{\todo}[1]{\textcolor{blue}{[TODO: #1]} }
\newcommand{\karl}[1]{\textcolor{brown}{$_{KS}$[#1]}}

\newcommand{\zewei}[1]{\textcolor{red}{$_{Z}$[#1]}}

\renewcommand{\todo}[1]{}
\renewcommand{\karl}[1]{}
\renewcommand{\zewei}[1]{}

\title{Mining Knowledge for Natural Language Inference \\from Wikipedia Categories}

\author{Mingda Chen$^{3}$\thanks{~~Equal contribution. Listed in alphabetical order.} \quad Zewei Chu$^{2*}$\quad Karl Stratos$^{1}$  \quad Kevin Gimpel$^{3}$ \\
$^{1}$Rutgers University, NJ, USA\\
$^{2}$University of Chicago, IL, USA\\
$^{3}$Toyota Technological Institute at Chicago, IL, USA\\
  \small\texttt{\{mchen,kgimpel\}@ttic.edu,zeweichu@gmail.com,stratos@cs.rutgers.edu}
}

\date{}
\date{}

\begin{document}
\maketitle
\begin{abstract}

Accurate lexical entailment (LE) and natural language inference (NLI) often require large quantities of costly annotations. To alleviate the need for labeled data, we introduce \wikinli: a resource for improving model performance on NLI and LE tasks. It contains 428,899 pairs of phrases constructed from naturally annotated category hierarchies in Wikipedia. We show that we can improve strong baselines such as \bert~\cite{devlin-etal-2019-bert} and \roberta~\cite{Liu2019roberta} by pretraining them on \wikinli and transferring the models on downstream tasks. We conduct systematic comparisons with phrases extracted from other knowledge bases such as \wordnet and \wikidata to find that pretraining on \wikinli gives the best performance.
In addition, we construct \wikinli in other languages, and show that pretraining on them improves performance on NLI tasks of corresponding languages.\footnote{Code and data are available at \href{https://github.com/ZeweiChu/WikiNLI}{\nolinkurl{https://github.com/ZeweiChu/WikiNLI}}.}
\end{abstract}

\section{Introduction}
Natural language inference (NLI) is the task of classifying the relationship, such as entailment or contradiction, between sentences.
It has been found useful in downstream tasks, such as summarization \cite{mehdad-etal-2013-abstractive} and long-form text generation \cite{holtzman-etal-2018-learning}.
NLI involves rich natural language understanding capabilities, many of which relate to world knowledge. To acquire such  knowledge, researchers have found benefit from external knowledge bases like \wordnet \cite{wordnet1998}, FrameNet \cite{baker-2014-framenet}, \wikidata \cite{wikidata2014}, and large-scale human-annotated datasets \cite{bowman-etal-2015-large,Williams2017mnli,nie2019adversarial}. Creating these resources generally requires expensive human annotation. In this work, we are interested in automatically generating a large-scale dataset from Wikipedia categories that can improve performance on both NLI and lexical entailment (LE) tasks.

One key component of NLI tasks is recognizing lexical and phrasal hypernym relationships. For example, vehicle is a hypernym of car. In this paper, we take advantage of the naturally-annotated Wikipedia category graph, where we observe that most of the parent-child category pairs are entailment relationships, i.e., a child category entails a parent category. %
Compared to \wordnet and \wikidata, the \wikipedia category graph has more fine-grained connections, which could be helpful for training models.
Inspired by this observation, we construct \wikinli, a dataset for training NLI models constructed automatically from the \wikipedia category graph, by automatic filtering from the \wikipedia category graph. The dataset has 428,899 pairs of phrases and contains three categories that correspond to the entailment and neutral relationships in NLI datasets.

To empirically demonstrate the usefulness of \wikinli, we pretrain \bert and \roberta on \wikinli, \wordnet, and \wikidata, before finetuning on various LE and NLI tasks. Our experimental results show that \wikinli gives the best performance averaging over 8 tasks for both \bert and \roberta.

We perform an in-depth analysis of approaches to handling the Wikipedia category graph and the effects of pretraining with \wikinli and other data sources under different configurations. We find that
\wikinli brings consistent improvements in a low resource NLI setting where there are limited amounts of training data, and the improvements plateau as the number of training instances increases; more \wikinli instances for pretraining are beneficial for downstream finetuning tasks with pretraining on a fourway variant of \wikinli showing more significant gains for the task requiring higher-level conceptual knowledge; \wikinli also introduces additional knowledge related to lexical relations benefiting finer-grained LE and NLI tasks.

We also construct \wikinli in other languages and benchmark several resources on XNLI \cite{conneau-etal-2018-xnli}, showing that \wikinli benefits performance on NLI tasks in the corresponding languages.

\section{Related Work}
\label{sec:related}

We build on a rich body of literature on leveraging specialized resources
(such as knowledge bases) to enhance model performance.
These works either (1) pretrain the model on datasets extracted from such resources,
or (2) use the resources directly by changing the model itself.

The first approach aims to improve performance at test time by designing
useful signals for pretraining, for instance using hyperlinks \cite{logeswaran-etal-2019-zero,chen-etal-2019-enteval}
or document structure in Wikipedia \cite{chen-etal-2019-evaluation},
knowledge bases \cite{logan-etal-2019-baracks}, and discourse markers \cite{nie-etal-2019-dissent}.
Here, we focus on using category hierarchies in Wikipedia.
There are some previous works that also use category relations derived from knowledge bases
\cite{shwartz-etal-2016-improving,riedel-etal-2013-relation}, but they are used in a particular form of distant supervision in which they are matched
with an additional corpus to create noisy labels.
In contrast, we use the category relations directly without requiring such additional steps. \citet{onoe2020fine} use the direct parent categories of hyperlinks for training entity linking systems.

Within this first approach, there have been many efforts aimed at harvesting inference rules from raw text~\citep{Lin:2001:DIR:973890.973894,szpektor-etal-2004-scaling,bhagat-etal-2007-ledir,szpektor-dagan-2008-learning,Yates:2009:UMD:1622716.1622724,bansal-etal-2014-structured,berant-etal-2015-efficient,hosseini2018learning}.
Since \wikinli uses category pairs in which one is a hyponym of the other, it is more closely related to work in extracting hyponym-hypernym pairs from text~\citep{hearst-1992-automatic,NIPS2004_2659,snow-etal-2006-semantic,pasca-07,mcnamee-etal-2008-learning,le-etal-2019-inferring}. \citet{pavlick2015adding} automatically generate a large-scale phrase pair dataset with several relationships by training classifiers on a relatively small amount of human-annotated data. However, most of this prior work uses raw text or raw text combined with either annotated data or curated resources like WordNet. \wikinli, on the other hand, seeks a middle road, striving to find large-scale, naturally-annotated data that can improve performance on NLI tasks.

The second approach aims to enable the model to leverage knowledge resources during prediction,
for instance by computing attention weights over lexical relations in WordNet \cite{chen2018neural}
or linking to reference entities in knowledge bases within the transformer block \cite{peters-etal-2019-knowledge}.
While effective, this approach requires nontrivial and domain-specific modifications of the model itself.
In contrast, we develop a simple pretraining method to leverage knowledge bases that
can likewise improve the performance of already strong baselines such as BERT
without requiring such complex model modifications.

There are some additional related works that focus on the category information of Wikipedia. \citet{Ponzetto2017derive} and \citet{Nastase2008decoding} extract knowledge of entities from the \wikipedia category graphs using predefined rules.
\citet{nastase-etal-2010-wikinet} build a dataset based on \wikipedia article
or category titles as well as the relations between categories and pages (``WikiNet''),
but they do not empirically validate the usefulness of the dataset.
In a similarly non-empirical vein, \citet{zesch-gurevych-2007-analysis} analyze the differences
between the graphs from \wordnet and the ones from \wikipedia categories.
Instead, we address the empirical benefits of leveraging the category information in the modern setting
of pretrained text representations.

\section{\wikinli}
\begin{figure}[t]
    \centering
    \includegraphics[scale=0.28]{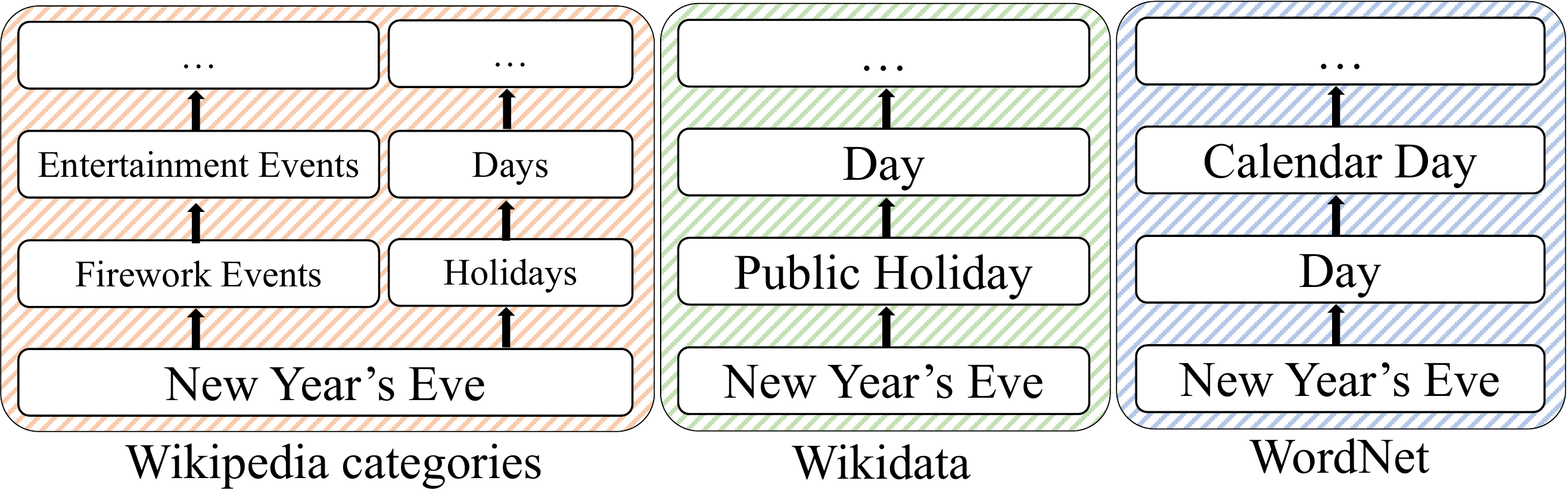}
    \caption{Example hierarchies obtained from \wikipedia categories, \wikidata, and \wordnet.
    }
    \label{fig:wikipedia_vs_wikidata_wordnet}
\end{figure}

We now describe how the \wikinli dataset is constructed from \wikipedia and its principal characteristics.
Each \wikipedia article is associated with crowd-sourced categories that correspond to topics or concepts covered by that article.
Wikipedia organizes these categories into a directed
graph that models their hierarchical relations.
For instance, the category ``Days'' is a parent node of
the category ``Holidays'' in this graph.
The central observation underlying \wikinli is that this
category hierarchy resembles the concept hierarchies and ontologies found in knowledge bases, such as \wikidata and \wordnet.

While there are similarities between the three resources, the \wikipedia category hierarchy contains more diverse connections between parent and child concepts. Figure~\ref{fig:wikipedia_vs_wikidata_wordnet} shows an example category ``New Year's Eve'' and its ancestors under these resources. All resources include a path that corresponds to the generalization of New Year's Eve as a regular day, but \wikipedia additionally includes a path that corresponds to the generalization as a celebration or entertainment event. Thus the \wikipedia hierarchy provides more abstract and fine-grained generalization that can be useful for NLI tasks. In this example, the commonsense knowledge that New Year's Eve implies entertainment is only directly captured by the \wikipedia hierarchy.

\wikinli is a dataset of category pairs extracted from this \wikipedia hierarchy to be used as an auxiliary task for pretraining NLI models. Specifically, \wikinli contains three types of category pairs based on their relations in the Wikipedia hierarchy: child-parent (``child''), parent-child (``parent''), and other pairs (``neutral''). The motivation is that child-parent resembles entailment; parent-child resembles reverse entailment; and other pairs resemble a neutral relationship. We find that this simple definition of relations is effective in practice; we also report an exploration with other types of relations such as siblings in experiments.

Table~\ref{tab:example_wikinli} shows examples from \wikinli that illustrate the diverse set of relations they address. They include conventional knowledge base entries such as ``Bone fractures'' being a type of ``Injuries'' and ``Chemical accident'' being a type of ``Pollution''. They also include relations that are more fine-grained than those typically found in knowledge bases. For instance, ``Pakistan'' is a child of ``South Asian countries''; in contrast, it is a child of ``Country'' in \wikidata. \wikinli also includes a large set of hyponym-hypernym relations in pairs that differ by only one or two words (e.g., ``Cantonese music'' and ``Cantonese culture''); their coverage is extensive and includes relations involving rare words such as ``Early Turkish Anatolia'' and ``Medieval Anatolia''.

\begin{table}[t]
\setlength{\tabcolsep}{5pt}
\small
\begin{center}
\begin{tabular}{|l|l|c|}
\hline
Category 1 & Category 2 & Rel. \\
\hline
Injuries  & Bone fractures & P \\
Chemical accident & Pollution & C \\
Armenian sportspeople & Curaçao male actors & N \\
Argentine design & Nigerian inventions & N \\
Cantonese music & Cantonese culture & C \\
Medieval Anatolia & Early Turkish Anatolia & P \\
Learned societies & Academic organizations & C\\
South Asian countries & Pakistan & P \\
\hline
\end{tabular}
\end{center}
\caption{\label{tab:example_wikinli}Examples from \wikinli. C = child; P = parent; N = neutral.}
\end{table}

More details of constructing \wikinli are as follows. We use the tables ``categorylinks'' and ``page'': these two pages provide category pairs in which one category is the parent of the other. We use all direct category relations. To eliminate trivial pairs, we remove pairs where either is a substring of the other. To construct neutral pairs, we randomly sample two categories where neither category is the ancestor of the other in the category graph. To make neutral pairs more ``related'' (so that they are harder to discriminate from direct relations), we encode both categories into continuous vectors using ELMo~\citep{peters2018deep} (averaging its three layers over all positions) and compute the cosine similarities between pairs.\footnote{We choose ELMo over BERT-like models because in our experiments, ELMo is better off-the-shelf without fine-tuning.} %
We pick the top-ranked pairs as neutral pairs in \wikinli. After the above processing, we remove categories longer than 50 characters and those containing certain keywords\footnote{all digits, ., !, ?, of, at, in, by, from, to, about, stubs, lists.} (see supplementary material for more results and examples on filtering criteria).
We ensure the dataset is balanced, and the final dataset has 428,899 unique pairs.

For the following experiments, unless otherwise specified, we only use 100,000 samples from \wikinli as training data and 5,000 as the development set since we find larger training set does not lead to performance gains (see Sec. \ref{sec:large_train_set} for more details). \wikinli is available at \url{https://github.com/ZeweiChu/WikiNLI}.

\section{Approach}
To demonstrate the effectiveness of \wikinli, we pretrain \bert and \roberta on \wikinli and other resources, and then finetune them
on several NLI and LE tasks. We assume that if a pretraining resource is better aligned with downstream tasks, it will lead to better downstream performance of the models pretrained on it.

\subsection{Training}
Following \citet{devlin-etal-2019-bert} and \citet{Liu2019roberta}, we use the concatenation of two texts as the input to \bert and \roberta.
Specifically, for a pair of input texts $x_1$, $x_2$, the input would be $\mathrm{[CLS]} x_1 \mathrm{[SEP]} x_2 \mathrm{[SEP]}$. We use the encoded representations at the position of $\mathrm{[CLS]}$ as the input to a two-layer classifier, and finetune the entire model.

We start with a pretrained \bertlarge or \robertalarge model and further pretrain it on different pretraining resources. After that, we finetune the model on the training sets for the downstream tasks, as we will elaborate on below.

\subsection{Evaluation}
We use several NLI and LE datasets. Statistics for these datasets are shown in Table~\ref{tab:dataset_stats} and details are provided below.

\begin{table}[t]
    \small
    \centering
    \begin{tabular}{|l|c|c|c|c|}
    \hline
        \textbf{dataset} & \textbf{\#train} & \textbf{\#dev} & \textbf{\#test} & \textbf{\#train per cat.} \\\hline
        \multicolumn{5}{|c|}{Natural Language Inference} \\\hline
        MNLI & 3,000 & 9,815 & 9,796 & 1,000 \\\hline
        SciTail & 3,000 & 1,304 & 2,126 & 1,500 \\\hline
        RTE & 2,490 & 277 & 3,000 & 1,245 \\\hline
        PPDB & 13,904 & 4,633 & 4,641 & 1,545 \\\hline
        Break & - & - & 8,193 & - \\\hline
        \multicolumn{5}{|c|}{Lexical Entailment} \\\hline
        K2010 & 739 & 82 & 621 & 370 \\\hline
        B2012 & 791 & 87 & 536 & 396 \\\hline
        T2014 & 539 & 59 & 507 & 270 \\\hline
    \end{tabular}
    \caption{Dataset statistics.}
    \label{tab:dataset_stats}
\end{table}

\subsubsection{Natural Language Inference}

\paragraph{MNLI.} The Multi-Genre Natural Language Inference (MNLI;~\citealp{Williams2017mnli}) dataset is a human-annotated multi-domain NLI dataset. MNLI has three categories: entailment, contradiction, and neutral. Since the training split for this dataset has a large number of instances, models trained on it are capable of picking up information needed regardless of the quality of the pretraining resources we compare, which makes the effects of pretraining resources negligible. To better compare
pretraining resources,
we simulate a low-resource scenario by randomly sampling 3,000 instances\footnote{The number of training instances is chosen based on the number of instances per category, as shown in the last column of Table \ref{tab:dataset_stats}, where we want %
the number to be close to 1-1.5K.} from the original training split as our new training set, but use the standard ``matched'' development and testing splits.

\paragraph{SciTail.} SciTail is created from science questions and the corresponding answer candidates, and premises from relevant web sentences retrieved from a large corpus \cite{scitail}. SciTail has two categories: entailment and neutral. Similar to MNLI, we randomly sample 3,000 instances from the training split as our training set.

\paragraph{RTE.}
We evaluate models on the GLUE \cite{wang2018glue} version of the recognizing textual entailment (RTE) dataset \cite{dagan2006pascal,bar2006second,giampiccolo2007third,bentivogli2009fifth}. RTE is a binary task, focusing on identifying if a pair of input sentences has the entailment relation.

\paragraph{PPDB.}

We use the human-annotated phrase pair dataset from \citet{pavlick2015adding}, which has 9 text pair relationship labels. The labels are: hyponym, hypernym, synonym, antonym, alternation, other-related, NA, independent, and none. We directly use phrases in PPDB to form input data pairs.
We include this dataset for more fine-grained evaluation.
Since there is no standard development or testing set for this dataset, we randomly sample 60\%/20\%/20\% as our train/dev/test sets.

\paragraph{Break.}

\citet{Glockner2018breaking} constructed a challenging NLI dataset called ``Break'' using external knowledge bases such as \wordnet. Since sentence pairs in the dataset only differ by one or two words, similar to a pair of adversarial examples, it has broken many NLI systems.

Due to the fact that Break does not have a training split, we use the aforementioned subsampled MNLI training set as a training set for this dataset.
We select the best performing model on the development set of MNLI and evaluate it on Break.

\subsubsection{Lexical Entailment}

We use the lexical splits for 3 datasets from \citet{levy-etal-2015-supervised}, including K2010 \cite{kotlerman-etal-2009-directional},
B2012 \cite{baroni-etal-2012-entailment}, and T2014 \cite{Turney2015ExperimentsWT}.
These datasets all similarly formulate lexical entailment as a binary task, and they were constructed from diverse sources, including human annotations, \wordnet, and \wikidata.

\section{Experiments}
\begin{table*}[t]
    \small
    \centering
    \begin{tabular}{l|c|c|c|c|c|c|c|c|c|}
        \cline{2-9}
        \multirow{2}*{} & \multicolumn{5}{|c|}{Natural Language Inference} & \multicolumn{3}{c|}{Lexical Entailment}\\ \cline{2-10}
        & MNLI & RTE & PPDB & Break & SciTail & K2010  & B2012 & T2014 & avg. \\
        \hline
        \multicolumn{1}{|l|}{\bert} & 75.0 & 69.9 & 66.7 & 80.2 &\underline{92.3} & \underline{85.2} &  79.4 & 63.3  & 76.5 \\
        \multicolumn{1}{|l|}{$+$\wordnet} & 75.8 & \underline{71.3} & 71.1 & 83.5 & 90.8 & 83.5 & 94.3 & \underline{71.2} & 80.2\\
        \multicolumn{1}{|l|}{$+$\wikidata} & 75.7 & \underline{71.3} & \bf\underline{75.0} & 81.3 & 91.5 & 82.3 & 95.3 & 70.5 & 80.4 \\
        \multicolumn{1}{|l|}{$+$\wikinli} & \underline{76.4} & 70.9 & 70.7 & \bf\underline{85.7}& 91.8  & 84.9 & \bf\underline{96.1}  & \underline{71.2} & \underline{81.0} \\
        \hline
        \multicolumn{1}{|l|}{\roberta} & 82.5 & 78.8 & 65.9 & 81.3 & 93.6 & 85.3 & 65.9 & 66.8 &  77.5 \\
        \multicolumn{1}{|l|}{$+$\wordnet} & 83.8 & 82.2 & 72.0 & 82.3 & \bf\underline{93.9}   & 82.5& 88.6  & 70.7 & 82.0 \\
        \multicolumn{1}{|l|}{$+$\wikidata} &  84.0 & 82.3 & \underline{72.5} & 83.2 & 92.9  & 82.4 & 94.8 & 71.0 & 82.9 \\
        \multicolumn{1}{|l|}{$+$\wikinli} & \bf\underline{84.4} &\bf \underline{83.1} & 71.7 & \underline{83.8}& 93.0 & \bf\underline{85.4}  & \underline{95.7} & \bf\underline{72.9} &\bf\underline{83.8} \\\hline
    \end{tabular}
    \caption{Test set performance for baselines and models %
    pretrained on various resources. We report accuracy (\%) for NLI tasks and $F_1$ score (\%) for LE tasks. The highest results for each model (\bert or \roberta) are underlined. The highest numbers in each column are boldfaced.
    }
    \label{tab:results}
\end{table*}

\subsection{Baselines}
We consider three baselines for both \bert and \roberta, namely the original model, the model pretrained on \wordnet, and the model pretrained on \wikidata.

\paragraph{\wordnet.} \wordnet is a widely-used lexical knowledge base, where words or phrases are connected by several lexical relations. We consider direct hyponym-hypernym relations available from \wordnet, resulting in 74,645 pairs.

\paragraph{\wikidata.} \wikidata is a database that stores items and relations between these items. Unlike \wordnet, \wikidata consists of items beyond word types and commonly seen phrases,
offering more diverse domains similar to \wikinli. The available conceptual relations in \wikidata are: ``subclass of'' and ``instance of''. In this work, we consider the ``subclass of'' relation in \wikidata because (1) it is the most similar relation to category hierarchies from Wikipedia; (2) the relation ``instance of'' typically involves more detailed information, which is found less useful empirically (see the supplementary material for details). The filtered data has 2,871,194 pairs.

We create training sets from both \wordnet and \wikidata following the same procedures used to create \wikinli. All three datasets are constructed from their corresponding parent-child relationship pairs. Neutral pairs are first randomly sampled from non-ancestor-descendant relationships and top ranked pairs according to cosine similarities of ELMo embeddings are kept. We also ensure these datasets are balanced among the three classes.

\subsection{Setup}

For all the experiments, we used the Hugging Face implementation~\cite{Wolf2019HuggingFacesTS}. When finetuning or pretraining~\bertlarge models, we mostly follow the hyperparameters suggested by \citet{devlin-etal-2019-bert}. Specifically, during pretraining, we use a batch size of 32, a learning rate of 2e-5, a maximum sequence length of 40, and 3 training epochs. During finetuning, we switch to use 8 as batch size due to memory constraints.
When funetuning or pretraining \robertalarge, we did extra hyperparameter search by adopting some of the hyperparameters recommended from~\citet{Liu2019roberta}.
We use 10\% training steps for learning rate warmup, 1e-5 for learning rate, and a maximum sequence length of 40, and train models for 3 epochs.\footnote{We choose this set of hyperparameters due to computational constraints. Our finetuned \roberta achieves 82.3\% accuracy on the RTE development set, which is lower than the 86.6\% accuracy reported in~\citet{Liu2019roberta}.}

For both models, we use development sets for model selection during pretraining. During downstream evaluations, we use a maximum sequence length of 128 for datasets involving sentences. We perform early stopping based on task-specific development sets and report the test results for the best models. Due to the variance of performance of 24-layer transformer architectures, we report medians of 5 runs with a fixed set of random seeds for all of our experiments. See the supplementary material for details on the runtime, hyperparameters, etc.

\subsection{Results}

The results are summarized in Table~\ref{tab:results}.
In general, pretraining on \wikinli, \wikidata, or \wordnet improves the performances on downstream tasks, and pretraining on \wikinli achieves the best performance on average. Especially for Break and MNLI, \wikinli can lead to much more substantial gains than the other two resources. Although \bertlarge + \wikinli is not better than the baseline \bertlarge on RTE, \roberta + \wikinli shows much better performance. Only on PPDB, \wikidata is consistently better than \wikinli. We note that \bertlarge + \wikinli still shows a sizeable improvement over the \bertlarge baseline.
More importantly, the improvements to both \bert and \roberta brought by \wikinli show that the benefit of the \wikinli dataset can generalize to different models. We also note that pretraining on these resources has little benefit for SciTail.

\section{Analysis}

We perform several kinds of analysis, including using \bert to compare the effects of different settings. Due to the submission constraints of the GLUE leaderboard, we will report dev set results (medians of 5 runs) for the tables in this section, except for Break which is only a test set.

\subsection{Lexical Analysis}

\begin{table}[t]
    \centering
    \small
\begin{tabular}{|c|c|c|}\hline
\wikinli & \wikidata & \wordnet \\\hline
albums & protein & genus \\
songs & gene & dicot \\
players & putative & family \\
male & protein-coding & unit \\
people & conserved & fish \\
American & hypothetical & tree \\
British & languages & bird \\
writers & disease & person \\
(band) & RNA & fern \\
\hline
\end{tabular}
    \caption{10 words from the top 20 most frequent words in \wikinli, \wikidata, and \wordnet. The full list is in the supplementary material.}
    \label{tab:lexical_analysis}
\end{table}
\begin{table}[t]
    \small\setlength{\tabcolsep}{4pt}
    \centering
    \begin{tabular}{l|c|c|c|c|c|}
        \cline{2-6}
        & MNLI & RTE & PPDB & Break  & avg. \\\hline
        \multicolumn{1}{|l|}{Threeway} & \bf 75.6 & \bf 74.4 & \bf 71.2 & 85.7 & \bf 76.7 \\
        \multicolumn{1}{|l|}{Fourway}  & \bf 75.6 & 74.0 & 69.8  & \bf 86.9 & 76.6 \\
        \multicolumn{1}{|l|}{Binary (C vs. R)}  & 75.1 & 72.6 & 70.5 & 81.7 & 75.0 \\
        \multicolumn{1}{|l|}{Binary (C/P vs. R)}  & 74.3 & 72.2 & 69.8 & 80.5 & 74.3  \\\hline
    \end{tabular}
    \caption{Comparing binary, threeway, and fourway classification for pretraining. C = child; P = parent; R = rest. The highest numbers in each column are boldfaced.}
    \label{tab:results_fourway}
\end{table}

\begin{table}[t]
    \small\setlength{\tabcolsep}{4pt}
    \centering
    \begin{tabular}{l|c|c|c|c|c|}
        \cline{2-6}
        & MNLI & RTE & PPDB & Break  & avg. \\\hline
        \multicolumn{1}{|l|}{Threeway 100k}  & 75.6 & 74.4 & \bf 71.2 & 85.7 & 76.7 \\
        \multicolumn{1}{|l|}{Fourway 100k}  & 75.6 & 74.0  & 69.8  & 86.9 & 76.6  \\
        \multicolumn{1}{|l|}{Threeway 400k} & \bf 75.7 & \bf 75.5 & 70.9 & 83.0
        & 76.3 \\
        \multicolumn{1}{|l|}{Fourway 400k} & 75.6 & 75.1 & 70.8 & \bf 89.5
        & \bf 77.8 \\
        \hline
    \end{tabular}
    \caption{The effect of the number of \wikinli pretraining instances. The highest numbers in each column are boldfaced.}
    \label{tab:results_full_dataset}
\end{table}
To qualitatively investigate the differences between \wikinli, \wikidata, and \wordnet, we list the top 20 most frequent words in these three resources in Table \ref{tab:lexical_analysis}. Interestingly, \wordnet contains mostly abstract words, such as ``unit'', ``family'', and ``person'', while \wikidata contains many domain-specific words, such as ``protein'' and ``gene''.
In contrast, \wikinli strikes a middle ground, covering topics broader than those in \wikidata but less generic than those in \wordnet.

\subsection{Fourway vs. Threeway vs. Binary Pretraining}

\begin{table}[t]
    \small\setlength{\tabcolsep}{4pt}
    \centering
    \begin{tabular}{l|c|c|c|c|c|}
        \cline{2-6}
        & MNLI & RTE & PPDB & Break  & avg. \\\hline
        \multicolumn{1}{|l|}{\TC{1} 100k}      & 75.6 & 74.4 & 71.2 & 85.7 & 76.7 \\
        \multicolumn{1}{|l|}{\TC{1} 50k}      & 74.9 & 74.7  & 70.8 & 76.9  & 74.3  \\
        \multicolumn{1}{|l|}{\TC{1} 50k + \TC{2} 50k} & 75.0 & 71.5 &  70.9 & 80.2 & 74.4 \\
        \multicolumn{1}{|l|}{\TC{1} 50k + \TC{3} 50k} & 75.0  &  73.6  & 70.7 & 81.5 &  75.3 \\
        \hline
    \end{tabular}
    \caption{Combining \wikinli with other datasets for pretraining. \TC{1}=\wikinli;\TC{2}=\wordnet;\TC{3}=\wikidata.}
    \label{tab:results_combine}
\end{table}

\begin{table*}[t]
    \small
    \centering
    \begin{tabular}{|l|l|c|c|c|c|c|}
    \hline
    phrase 1 & phrase 2 & gold & \bert & \wikinli & \wordnet & \wikidata \\ \hline
        car & the trunk & hypernym & other-related & hypernym & hypernym & hypernym \\
        return & return home & hypernym & synonym & hypernym & hypernym & hypernym \\
        boys are & the children are & hyponym & synonym & hyponym & hyponym & hyponym \\
        foreign affairs & foreign minister & other-related & hypernym & other-related & hypernym & hypernym \\
        company & debt & other-related & independent & independent & other-related & other-related \\
        europe & japan & alternation & hypernym & alternation & independent & alternation\\
        family & woman & independent & independent & hypernym & independent & other-related\\\hline
    \end{tabular}
    \caption{Examples from PPDB development set showing the effect of pretraining resources.}
    \label{tab:effect_pretrain}
\end{table*}

\begin{table}[t]
    \small
    \centering
    \begin{tabular}{l|c|c|c|c|c|}
        \cline{2-6}
        & 2k & 3k & 5k & 10k & 20k  \\
        \hline
        \multicolumn{6}{|c|}{MNLI}\\\hline
        \multicolumn{1}{|l|}{\bert} & 72.2  & 74.4 & 76.6 & 78.8 & 80.4  \\
        \multicolumn{1}{|l|}{\wikinli} & 74.5 & 75.6 & 77.3  & 79.1 & 80.6 \\\hline
        \multicolumn{1}{|c|}{$\Delta$} & +2.3 & +1.2 & +0.7  & +0.3 & +0.2 \\\hline
        \multicolumn{6}{|c|}{PPDB}\\ \hline
        \multicolumn{1}{|l|}{\bert} & 55.5 & 59.2 & 59.9  & 68.1 & 68.6 \\
        \multicolumn{1}{|l|}{\wikinli} & 65.0 & 66.4 & 67.9  & 70.2 & 71.2 \\\hline
        \multicolumn{1}{|c|}{$\Delta$} & +9.5 & +7.2 & +8.0 & +2.1 & +2.6 \\\hline
    \end{tabular}
    \caption{Results for varying numbers of MNLI or PPDB training instances. The rows ``$\Delta$'' show improvements from \wikinli. We use all the training instances for PPDB in the ``20k'' setting.
    }
    \label{tab:results_mnli}
\end{table}

\begin{table}[t]
    \small
    \centering
    \begin{tabular}{c|c|c|c|c|}\cline{2-5}
                 & antonym & alternation  & hyponym & hypernym \\ \hline
        \multicolumn{1}{|l|}{w/ } & 34& 51 & 276 & 346  \\
        \multicolumn{1}{|l|}{w/o}& 1 & 35 & 231 & 248 \\\hline
    \end{tabular}
    \caption{Per category numbers of correctly predicted instances by \bert with or without pretraining on \wikinli.\protect\footnotemark }
    \vspace{-1em}
    \label{tab:effect_category}
\end{table}

\begin{table}[t]
    \centering
    \small
    \begin{tabular}{|c|c|c|c|}\hline
& R1 & R2 & R3 \\\hline
BERT & 39.8 & 37.0 & 41.3 \\
$+$ \wordnet & 41.1 & 38.2 & 39.9 \\
$+$ \wikidata & 43.2 & 39.0 & 41.8 \\
$+$ \wikinli & 39.6 & 38.2 & 39.3 \\\hline
RoBERTa & 46.1 & 39.3 & 39.4 \\
$+$ \wordnet & 53.7 & 38.7 & 37.9 \\
$+$ \wikidata & 51.5 & 39.6 & 39.8 \\
$+$ \wikinli & 51.2 & 38.1 & 39.4 \\\hline
\end{tabular}
    \caption{Test results for ANLI.}
    \label{tab:anli}
\end{table}

We investigate the effects of the number of categories for \wikinli by empirically comparing three settings: fourway, threeway, and binary classification.
For fourway classification, we add an extra relation ``sibling'' in addition to child, parent, and neutral relationships. A sibling pair consists of two categories that share the same parent. We also ensure that neutral pairs are non-siblings, meaning that we separate a category that was considered as part of the neutral relations to provide a more fine-grained pretraining signal.

We construct two versions of \wikinli with binary class labels. One classifies the child against the rest, including parent, neutral, and sibling (``child vs. rest''). The other classifies child or parent against neutral or sibling (``child/parent vs. rest''). The purpose of these two datasets is to find if a more coarse training signal would reduce the gains from pretraining. %

These dataset variations are each balanced among their classes and contain 100,000 training instances and 5,000 development instances.

Table~\ref{tab:results_fourway} shows results
of MNLI, RTE, and PPDB.
Overall, fourway and threeway classifications are comparable, although they excel at different tasks. Interestingly, we find that pretraining with  child/parent vs.~rest is worse than pretraining with child vs.~rest. We suspect this is because the child/parent vs.~rest task resembles topic classification. The model does not need to determine direction of entailment, but only whether the two phrases are topically related, as neutral pairs are generally either highly unrelated or only vaguely related.
The child vs.~rest task still requires reasoning about entailment as the models still need to differentiate between child and parent.

In addition, we explore pruning levels in Wikipedia category graphs, and incorporating sentential context, finding that relatively higher levels of knowledge from \wikinli have more potential of enhancing the performance of NLI systems and sentential context shows promising results on the Break dataset (see supplementary material for more details).

\subsection{Larger Training Sets}
\label{sec:large_train_set}

We train on larger numbers of \wikinli instances, approximately 400,000, for both threeway and fourway classification. We note that we only pretrain models on \wikinli for one epoch as it leads to better performance on downstream tasks. The results are in Table~\ref{tab:results_full_dataset}. We observe that except for PPDB, adding more data generally improves performance. For Break, we observe significant improvements when using fourway \wikinli for pretraining, whereas threeway \wikinli seems to hurt the performance.

\subsection{Combining Multiple Data Sources}

We combine multiple data sources for pretraining. In one setting we combine 50k instances of \wikinli with 50k instances of \wordnet, while in the other setting we combine 50k instances of \wikinli with 50k instances of \wikidata. Table~\ref{tab:results_combine} compares these two settings for pretraining. \wikinli works the best when pretrained alone.

\footnotetext{We observed similar trends when pretraining on the other resources.}

\subsection{Effect of Pretraining Resources}

We show several examples of predictions from PPDB in Table~\ref{tab:effect_pretrain}. In general, we observe that without pretraining, \bert tends to predict symmetric categories, such as synonym, or other-related, instead of predicting entailment-related categories. For example, the phrase pair ``car'' and ``the trunk'', ``return'' and ``return home'', and ``boys are'' and ``the children are''. These are either ``hypernym'' or ``hyponym'' relationship, but \bert tends to conflate them with symmetric relationships, such as other-related. To quantify this hypothesis, we compute the numbers of correctly predicted antonym, alternation, hyponym and hypernym and show them in Table~\ref{tab:effect_category}. It can be seen that with pretraining those numbers increase dramatically, showing the benefit of pretraining on these resources.

We also observe that the model performance can be affected by the coverage of pretraining resources. In particular, for phrase pair ``foreign affairs'' and ``foreign minister'',
\wikinli has a closely related term ``foreign affair ministries'' and ``foreign minister'' under the category ``international relations'', whereas \wordnet does not have these two, and \wikidata only has ``foreign minister''.

As another example, consider the phrase pair ``company'' and ``debt''. In \wikinli, ``company'' is under the ``business'' category and debt is under the ``finance'' category. They are not directly related. In \wordnet, due to the polysemy of ``company'', ``company'' and ``debt'' are both hyponyms of %
``state'', and in \wikidata, they are both a subclass of ``legal concept''.

For the phrase pair ``family''/``woman'',
in \wikinli, ``family'' is a parent category of ``wives'', and in \wikidata, they are related in that ``family'' is a subclass of ``group of humans''. In contrast, \wordnet does not have such knowledge.

\subsection{Finetuning with Different Amounts of Data}

We now look into the relationship between the benefit of \wikinli and the number of training instances from downstream tasks (Table \ref{tab:results_mnli}). We compare \bertlarge to \bertlarge pretrained on \wikinli when finetuning on 2k, 3k, 5k, 10k, and 20k MNLI or PPDB training instances accordingly. In general, the results show that \wikinli has more significant improvement with less training data, and the gap between \bertlarge and \wikinli narrows as the training data size increases. We hypothesize that the performance gap does not reduce as expected between 3k and 5k or 10k and 20k due in part to the imbalanced number of instances available for the categories. For example, even when using 20k training instances, some of the PPDB categories are still quite rare.

\subsection{Evaluating on Adversarial NLI}

Adversarial NLI (ANLI; \citealp{nie2019adversarial}) is collected via an iterative human-and-model-in-the-loop procedure. ANLI has three rounds that progressively increase the difficulty. When finetuning the models for each round, we use the sampled 3k instances from the corresponding training set, perform early stopping on the original development sets, and report results on the original test sets. As shown in Table \ref{tab:anli}, our pretraining approach has diminishing effect as the round number increases. This may due to the fact that humans deem the NLI instances that require world knowledge as the hard ones, and therefore when the round number increases, the training set is likely to have more such instances, which makes pretraining on similar resources less helpful.
Table \ref{tab:anli} also shows that \wordnet and \wikidata show stronger performance than \wikinli. We hypothesize that this is because ANLI has a context length almost 3 times longer than MNLI on average, in which case our phrase-based resources or pretraining approach are not optimal choices. Future research may focus on finding better ways to incorporate sentential context into \wikinli. For example, we experiment with such a variant of \wikinli (i.e., \textsc{WikiSentNLI}) in the supplementary material.

We have similar observations that our phrase-based pretraining has complicated effect (e.g., only part of the implicature results shows improvements) when evaluating these resources on IMPPRES \cite{jeretic-etal-2020-natural}, which focuses on the information implied in the sentential context (please refer to the supplementary materials for more details).

\section{Multilingual \wikinli}

\begin{table}[t]
    \small\setlength{\tabcolsep}{6pt}
    \centering
    \begin{tabular}{l|c|c|c|c|c|}
        \cline{2-6}
        & fr & ar & ur & zh & avg. \\\hline
        \multicolumn{1}{|l|}{mBERT}      & 61.5  & 57.3  & 49.3 & 57.9 & 56.5 \\
        \multicolumn{1}{|l|}{m\wikinli}  & 62.5 & 56.8 & 51.5 & 59.9 & 57.7 \\
        \multicolumn{1}{|l|}{tr\wikinli}  & 63.0 & \bf 57.7  & 51.3  &  59.9  & 58.0 \\
        \multicolumn{1}{|l|}{\wikinli}  & \bf 63.3 & 57.1 & \bf 51.8  &\bf 60.0 & \bf 58.1 \\
        \multicolumn{1}{|l|}{\wikidata} & 63.2 & 56.9 &  49.5 & 59.8 & 57.4 \\
        \multicolumn{1}{|l|}{\wordnet}  & 63.1 & 56.0 & 50.5 & 58.6 & 57.1 \\\hline
    \end{tabular}
    \caption{Test set results for XNLI. m\wikinli is constructed from Wikipedia in other languages. tr\wikinli is translated from the English \wikinli. The highest numbers in each column are boldfaced.}
    \vspace{-0.5em}
    \label{tab:results_other_langs}
\end{table}

Wikipedia has different languages, which naturally motivates us to extend \wikinli to other languages.
We mostly follow the same procedures as English \wikinli to construct a multilingual version of \wikinli from Wikipedia in other languages, except that (1) we filter out instances that contain English words for Arabic, Urdu, and Chinese; and (2) we translate the keywords into Chinese when filtering the Chinese \wikinli.
We will refer to this version of \wikinli as ``m\wikinli''. As a baseline, we also consider ``tr\wikinli'', where we translate the English \wikinli into other languages using Google Translate. We benchmark these resources on XNLI in four languages: French (fr), Arabic (ar), Urdu (ur), and Chinese (zh).
When reporting these results, we pretrain multilingual BERT (mBERT; \citealp{devlin-etal-2019-bert}) on the corresponding resources, finetune it on 3000 instances of the training set, perform early stopping on the development set, and test it on the test set. We always use XNLI from the corresponding language. In addition, we pretrain mBERT on English \wikinli, \wikidata, and \wordnet, finetune and evaluate them on other languages using the same language-specific 3000 NLI pairs mentioned earlier.
We note that when pretraining on m\wikinli or tr\wikinli, we use the versions of these datasets with the same languages as the test sets.

Table~\ref{tab:results_other_langs} summarizes the test results on XNLI. In general, pretraining on \wikinli gives the best results. \citet{phang2020english} also observed that training on English intermediate tasks helps in cross-lingual tasks but in a zero-shot setting. While m\wikinli is not the best resource, it still gives better results on average than \wikidata, \wordnet, and no pretraining at all. The  exception is Arabic, where only tr\wikinli performs better than the mBERT baseline. In comparing among different versions of \wikinli, we find that tr\wikinli  performs almost as good as \wikinli, but for Urdu, tr\wikinli is the worst resource among the three.
The variance of tr\wikinli may arise from the variable quality of machine translation across languages.

The accuracy differences between  m\wikinli and \wikinli could be partly attributed to domain differences across languages. To measure the  differences, we compile a list of the top 20 most frequent words in the Chinese m\wikinli, shown in Table \ref{tab:en_vs_zh_word_list}. The most frequent words for m\wikinli in Chinese are mostly related to political concepts, whereas \wikinli offers a broader range of topics.

Future research will be required to obtain a richer understanding of how training on \wikinli benefits non-English languages more than training on the language-specific m\wikinli resources. One possibility is the presence of emergent cross-lingual structure in mBERT~\citep{wu2019emerging}. Nonetheless, we believe m\wikinli and our training setup offer a useful framework for further research into multilingual learning with pretrained models.

\begin{CJK*}{UTF8}{gbsn}
\begin{table}[t]
    \centering
    \small
\begin{tabular}{|c|c|}\hline
\wikinli & Chinese m\wikinli \\\hline
albums & 中国 (China) \\
songs & 中华人民共和国 (P. R. C.) \\
players & 行政区划 (administrative division) \\
male & 人 (man) \\
people & 政治 (politics) \\
American & 人物 (people) \\
British & 各国 (countries) \\
writers & 组织 (organization) \\
(band) & 各省 (provinces) \\
female & 建筑物 (building) \\
\hline
\end{tabular}

    \caption{10 words from the top 20 most frequent words in \wikinli, and m\wikinli in Chinese. Each Chinese word is followed by a translation in parenthesis. The full list is shown in the supplementary material.
    }
    \vspace{-1em}
    \label{tab:en_vs_zh_word_list}
\end{table}
\end{CJK*}

\section{Conclusion}

We constructed \wikinli, a large-scale naturally-annotated dataset for improving model performance on NLI and LE tasks. Empirically, we benchmarked \wordnet, \wikidata, and \wikinli using both \bert and \roberta by first pretraining these models on those resources, then finetuning on downstream tasks. The results showed that pretraining on \wikinli gives the largest gains averaging over 8 different datasets. The improvements to both \bert and \roberta showed that the benefit of \wikinli can generalize. We also performed an in-depth analysis on ways of constructing \wikinli, and a lexical analysis on the differences between the three benchmarked resources. Our experiments on m\wikinli showed promising results and can benefit the research on multilinguality.

\section*{Acknowledgments}
This research was supported in part by a Bloomberg data science research grant to K.~Stratos and K.~Gimpel.

\bibliography{anthology,acl2020}
\bibliographystyle{acl_natbib}

\end{document}